\documentclass[runningheads]{llncs}
\usepackage{graphicx}
\usepackage{url}
\usepackage{mathrsfs}
\usepackage{setspace}
\usepackage{amsmath,amssymb} 
\usepackage{color}
\usepackage[width=122mm,left=12mm,paperwidth=146mm,height=193mm,top=12mm,paperheight=217mm]{geometry}
\begin{document}
\pagestyle{headings}
\mainmatter

\def\ECCV16SubNumber{1294}  

\title{Dual Purpose Hashing} 
\author{Haomiao Liu$^{1,2}$, Ruiping Wang$^1$, Shiguang Shan$^1$, Xilin Chen$^1$}
\institute{$^1$Key Lab of Intelligent Information Processing of Chinese Academy of Sciences(CAS),\\
Institute of Computing Technology, CAS, Beijing, 100190, China\\
$^2$University of Chinese Academy of Sciences, Beijing, 100049, China\\
{\tt\small haomiao.liu@vipl.ict.ac.cn, \{wangruiping, sgshan, xlchen\}@ict.ac.cn}}

\maketitle

\vspace{-0.5cm}

\begin{abstract}

Recent years have seen more and more demand for a unified framework to address multiple realistic image retrieval tasks concerning both category and attributes. Considering the scale of modern datasets, hashing is favorable for its low complexity. However, most existing hashing methods are designed to preserve one single kind of similarity, thus improper for dealing with the different tasks simultaneously. To overcome this limitation, we propose a new hashing method, named Dual Purpose Hashing (DPH), which jointly preserves the category and attribute similarities by exploiting the Convolutional Neural Network (CNN) models to hierarchically capture the correlations between category and attributes. Since images with both category and attribute labels are scarce, our method is designed to take the abundant partially labelled images on the Internet as training inputs. With such a framework, the binary codes of new-coming images can be readily obtained by quantizing the network outputs of a binary-like layer, and the attributes can be recovered from the codes easily. Experiments on two large-scale datasets show that our dual purpose hash codes can achieve comparable or even better performance than those state-of-the-art methods specifically designed for each individual retrieval task, while being more compact than the compared methods.

\end{abstract}

\begin{spacing}{0.976}

\section{Introduction}\label{sec:intro}

In recent years, more and more images are available on the Internet, posing great challenges to retrieving images relevant to a given query image. At the meantime, the retrieval tasks have also become more diverse. In real-life scenarios, three common retrieval tasks are: {\bf\uppercase\expandafter{\romannumeral1}}. retrieving images from the same category as the query image, e.g. matching street clothing photos in online shops \cite{hadi2015buy}; {\bf\uppercase\expandafter{\romannumeral2}}. retrieving images with specified attributes, e.g. looking for young Asian woman wearing sunglasses \cite{siddiquie2011image}; and {\bf\uppercase\expandafter{\romannumeral3}}. the combination of the above tasks, e.g. looking for clothing of the same style but with a different color. Existing algorithms \cite{hadi2015buy,siddiquie2011image,babenko2014neural,yu2012weak} can be adopted to tackle the above tasks, and have achieved certain degree of successes. However, the high complexities of indexing and retrieving with real-valued image representations limit the scalability of such methods. To deal with this problem, hashing is often adopted for its high efficiency in both time and storage.

\begin{figure}[t]
\setlength{\belowcaptionskip}{-0.5 cm}
\begin{center}
\begin{tabular}{ccc}
\includegraphics[width=4.0 cm]{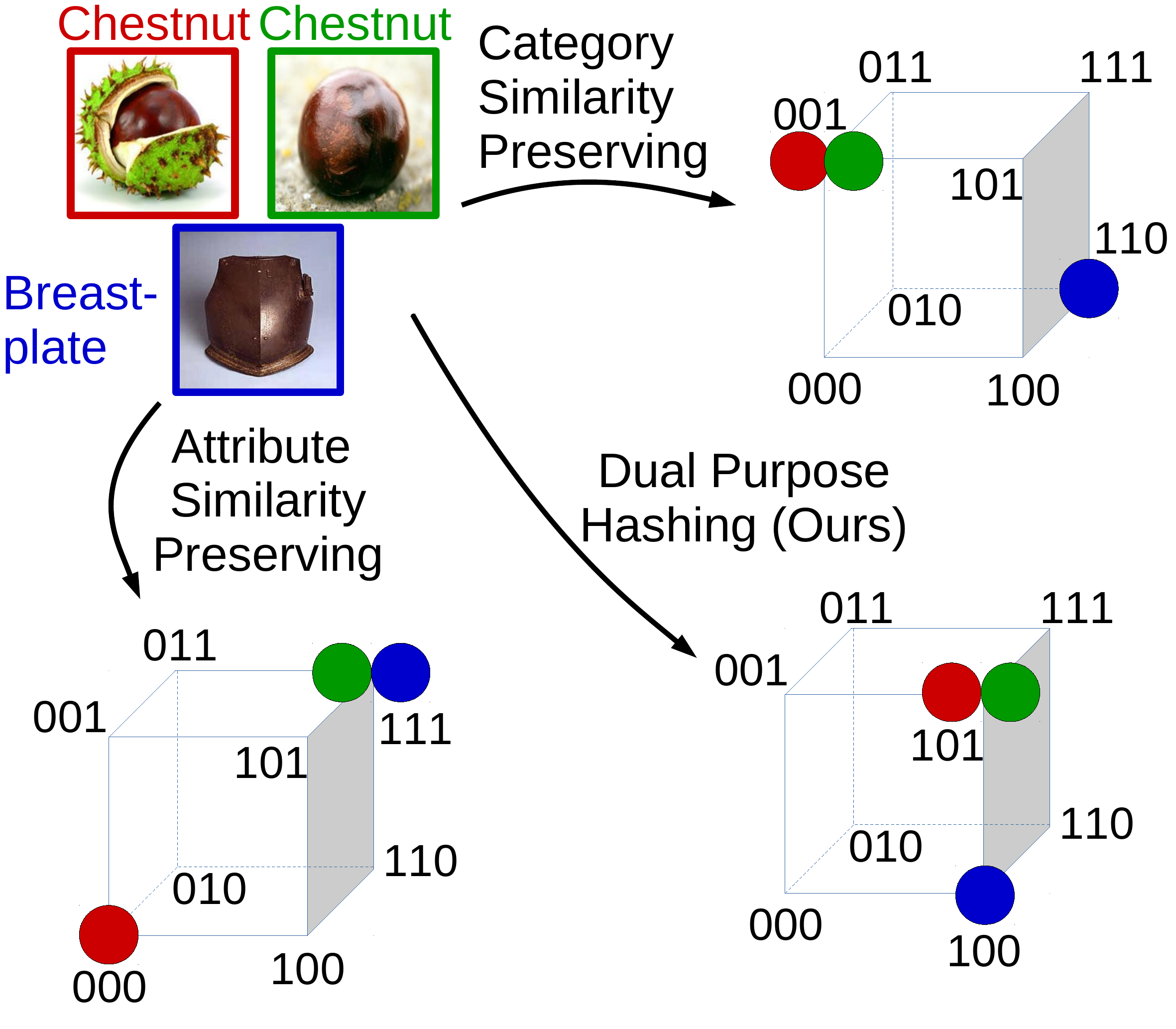} &  
~~~~~~~&
\includegraphics[width=7.0 cm]{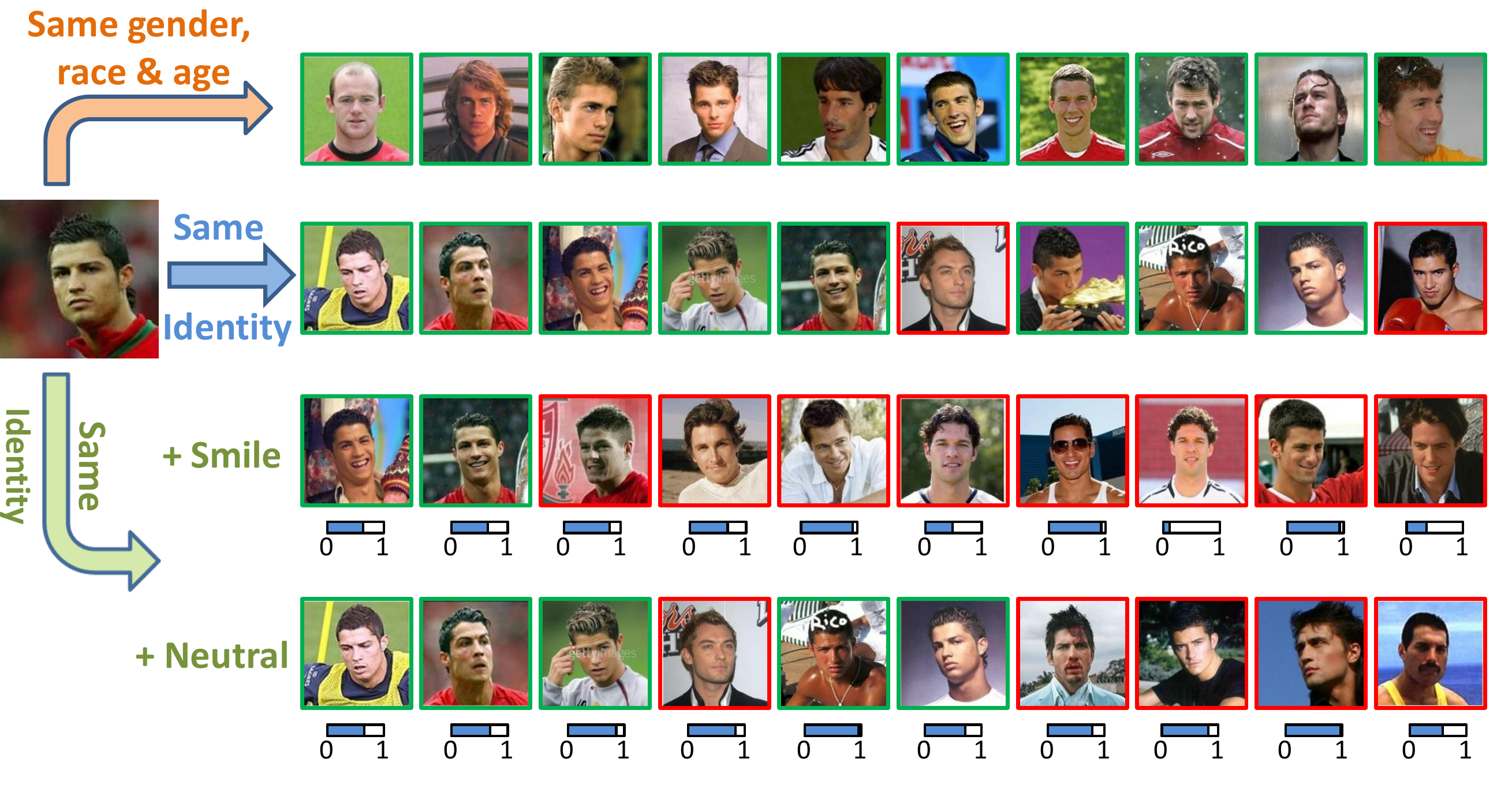} \\
(a) & & (b)
\end{tabular}
\end{center}
\vspace{-0.5 cm}
\caption{(a)Illustration of the idea of our Dual Purpose Hashing method. (b) A real example showing the three retrieval tasks on a face dataset, where the query is an image of Cristiano Ronaldo. The top ranked feedbacks of each task are shown here. In the first two rows, images exactly match the query are bounded by green boxes, and red otherwise. In the last two rows, images of the same/different identity are bounded by green/red boxes respectively, and the blue bars below the images indicate the confidence level of the corresponding attribute. Best viewed on a computer screen.}
\label{fig:illustration}
\end{figure}

A major issue concerning most existing hashing methods is that they are usually designed to preserve one single kind of similarity, e.g. semantic similarity defined by categories. Due to the difference between attributes and category, multiple models would be needed to preserve both category and attribute similarities for satisfactory performance. However, such scheme is suboptimal since training multiple models is time-consuming, and the redundancies between the models might harm the storage efficiency. To tackle this issue, we propose a unified framework to jointly preserve both similarities, named Dual Purpose Hashing (DPH), as illustrated in Figure \ref{fig:illustration}(a). In our DPH method, only a single model is learned to produce binary codes that can be used to simultaneously deal with the three tasks above, thus reducing the training time and redundancies in storage. Figure \ref{fig:illustration}(b) shows a real face image retrieval case of our method on a challenging face dataset.

Our basic idea comes from a very natural intuition that category and attributes, as objects' descriptions at different semantic levels, should share some common low-level visual features. This can be partly confirmed from the experimental studies in some recent works \cite{escorcia2015relationship,zhong2016face}, where it is shown that some nodes in the top layers of CNNs trained for classification tasks are highly correlated with visual attributes. Such observations also suggest that deep CNN model is a good choice to hierarchically capture the correlations between category and attributes. This motivates us to adopt CNN models to learn unified binary codes that can preserve both similarities simultaneously.

\begin{figure}[t]
\setlength{\belowcaptionskip}{-0.8 cm}
\begin{center}
\includegraphics[width=12.2 cm]{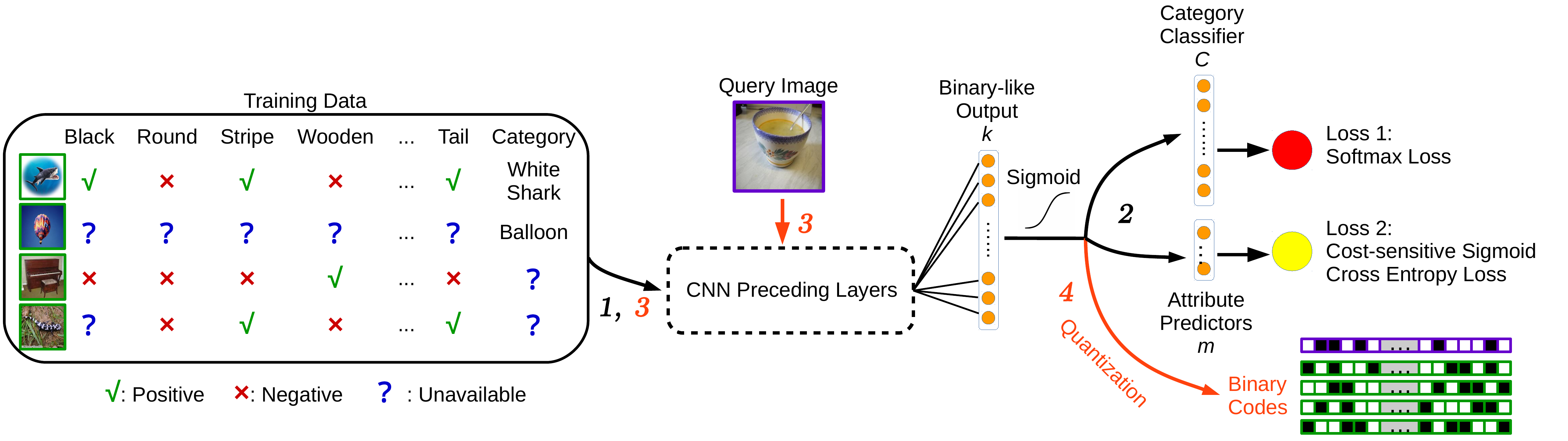}
\caption{The framework of our proposed DPH method. To simultaneously encode category and visual attributes of images into binary codes, we devise a CNN model that can take partially labelled images as training input (step 1), and train the model on classification and attribute prediction tasks (step 2). The binary-like output layer, which have $k$ (the code length) nodes, is connected to the two task layers as input. To produce binary codes, images are propagated through the network (step 3), and the binary-like network output is quantized (step 4).}
\label{fig:framework}
\end{center}
\end{figure}

The framework of our DPH method is illustrated in Figure \ref{fig:framework}. To be specific, our network contains a binary-like layer, which is used to approximate the binary code. By jointly optimizing a classification loss and an attribute prediction loss, our method can encode both similarities into the binary codes. Since most images available on the Internet do not have complete category and attribute labels, our loss function is properly designed to take into account such practical scenarios, namely, even images with only one label can contribute to the model learning. By doing so, an additional benefit is that the network has the capacity to see a large amount of partially labelled data in the training stage, and thus greatly reduces the risk of overfitting. 

Once the model is learned, images can be indexed by quantizing the outputs of the binary-like layer to compact hash codes. In the first task relevant to category, retrieval can be done similarly to existing hashing methods by utilizing Hamming distance ranking or hash table lookup. For the last two tasks relevant to attributes, real-valued attribute predictions (in general, such real-valued predictions are more powerful than binary-valued alternatives) can be recovered from the binary codes with a simple matrix multiplication operation, which can be efficiently done by only a few summation operations. Compared with directly storing the real-valued attribute predictions, our method only incurs a little increase in computation cost, while dramatically reduces the storage space.

The main contributions of this work are two-fold: First, we present a unified framework to learn hash functions that simultaneously preserve category and attribute similarities for addressing multiple retrieval tasks. Second, we propose a new training scheme for the CNN models that can take partially labelled data as training inputs to improve the performance and alleviate overfitting.


\section{Related Works}\label{sec:relwork}

In large-scale retrieval tasks, hashing \cite{gionis1999similarity,weiss2009spectral,kulis2009learning,wang2012semi,gong2011iterative,norouzi2011minimal,liu2012supervised,xia2014supervised,lai2015dnnh} is favorable for its low time and space complexity. The pioneering data-independent hashing method Locality Sensitive Hashing (LSH) \cite{gionis1999similarity}, uses random projections to produce binary bits. However, LSH usually requires long codes to achieve satisfactory retrieval performance, which demands for large amount of storage space.

To reduce the storage cost, data-dependent hashing methods are proposed to learn more compact binary codes by utilizing a training set. Such methods can be further divided into unsupervised and supervised (semi-supervised). Unsupervised methods only make use of unlabelled training data to learn hash functions. For example, Spectral Hashing (SH) \cite{weiss2009spectral} aims at minimizing the weighted Hamming distance of binary codes, where the weights are defined as the similarity metrics of image pairs; Iterative Quantization (ITQ) \cite{gong2011iterative} attempts to orthogonally transform the image descriptor such that the quantization errors are minimized.

Supervised methods are proposed to deal with the more complicated semantic similarities by taking advantage of semantic labels. CCA-ITQ \cite{gong2011iterative} boosts the performance of ITQ by using label information to obtain more discriminative image descriptors; Minimal Loss Hashing (MLH) optimizes an upper bound
of a hinge-like loss to learn the hash functions. On the other hand, Semi-Supervised Hashing (SSH) \cite{wang2012semi} uses unlabelled data to improve the generalization ability of the hash functions. Since the above methods use linear hash functions, they can hardly deal with linearly inseparable data. To overcome this limitation, Binary Reconstructive Embedding (BRE) \cite{kulis2009learning} and Supervised Hashing with Kernels (KSH) \cite{liu2012supervised} are proposed to take advantage of the non-linearity of kernel spaces.

Although the aforementioned hashing methods have achieved successes in some applications, they use the readily extracted features, which are not specifically designed for the task at hand, thus might lose some task-specific information. To tackle this issue, most recently, several hashing methods \cite{zhao2015deep,lai2015dnnh,xia2014supervised,lin2015deep,zhang2015bitscalable} significantly improve the state-of-the-arts by jointly learning the image representations and the hash functions using CNN models.

Besides category label-oriented image retrieval, attributes have been widely adopted to deal with some other realistic retrieval tasks \cite{sadovnik2013spoken,kovashka2013attribute,scheirer2012multi,kumar2008facetracer,tao2015attributes,turakhia2013attribute,rastegari2013multi,yu2012weak,siddiquie2011image}. This paper is most related to the works that use nameable attributes \cite{parikh2011interactively} as queries. \cite{kumar2008facetracer} predicts the probability of attributes with SVM classifiers, and uses the product of probabilities to rank the database images. Follow-up works investigate the usage of attribute correlation \cite{siddiquie2011image}, fusion strategy \cite{scheirer2012multi,rastegari2013multi}, relative attributes \cite{sadovnik2013spoken}, and other techniques \cite{kovashka2013attribute,turakhia2013attribute} to improve the retrieval performance. In this paper, we adopt the retrieval strategy in \cite{kumar2008facetracer} for simplicity, while those more complicated ones \cite{siddiquie2011image,scheirer2012multi,rastegari2013multi} are also compatible with our framework. A major issue of these attributes-oriented image retrieval methods is the usage of real-valued features, which limits the scalability and efficiency of such methods.

In light of the successes of hashing methods, recently \cite{rastegari2012attribute,li2015two} have made some early attempts to connect attributes with binary codes. \cite{rastegari2012attribute} tries to discover attributes after the hash functions are learned by visualizing the images with the highest and lowest scores at each bit. This ``post-processing'' manner, however, hinders the method to learn the desired nameable attributes, thus making \cite{rastegari2012attribute} unsuitable to be used for attribute-oriented retrieval tasks. \cite{li2015two} improves \cite{rastegari2012attribute} by explicitly modeling the connection between hash bits and attributes in the binary code learning stage. Nevertheless, the simple linear transformation based on the manually selected image representations in \cite{li2015two} is obviously inadequate to capture the complex correlation between category and attributes. To address the shortcomings of previous works, we propose to exploit the CNN models to hierarchically extract the correlation between these two semantic descriptions in an end-to-end manner.



\section{Approach}\label{sec:app}

Our goal is to learn compact binary codes such that: a) images from the same category are encoded to similar binary codes; b) images with similar attributes should have similar binary codes; c) the learned model should generalize well to new-coming images.

To achieve this goal, we present a hash learning framework as illustrated in Figure \ref{fig:framework}. The preceding layers of the network consists of several convolution-pooling layers, and optionally followed by several fully connected layers. The structure of these layers is very flexible, thus various successful models \cite{krizhevsky2012imagenet,szegedy2015going,he2015deep} can be adopted in our method. Since directly optimizing binary codes is difficult, the penultimate layer in our network is designed to give binary-like outputs (a fully connected layer with $sigmoid$ activation) to approximate the binary codes. During the training stage, the whole network is jointly trained on classification and attribute prediction tasks to encode both kinds of semantic information into binary codes. Moreover, the loss functions are specifically designed to make use of the abundant partially labelled data on the Internet, which can meanwhile improve the generalization ability of the models, as shown in Section \ref{sec:exp:module}.

\subsection{Problem Setup}

Let $\Omega$ be the space of RGB images, we want to train an end-to-end model that maps images from $\Omega$ to $k$-bit binary codes $\mathscr{F}: \Omega \rightarrow \{0, 1\}^k$. Suppose that the training images are from $C$ known categories, and annotated with a set of $m$ visual attributes. Let $S_{tr}=\{(X_i^{tr},y_i,{\bf a}_i)|i=1,\cdots,N\}$ denote the training set consisting of $N$ images, where $X_i^{tr}\in\Omega$, $y_i\in\{1,\cdots,C,C+1\}$ is the category label of the $i$-th image, and ${\bf a}_i\in\{0,1,2\}^m$ are the visual attribute labels. More specifically, $y_i=C+1$ means the category label of the $i$-th image is missing. ${\bf a}_{ij}=1$ and $0$ indicates the $j$-th attribute is present/absent in the $i$-th image. Besides, we use ${\bf a}_{ij}=2$ to denote that the $j$-th attribute label of the $i$-th image is missing. Each training image is required to have at least one available label.

\subsection{Category Information Encoding}\label{sec:app:cls}

To preserve category similarity, our basic idea is that if a simple transformation (e.g. softmax classifier) can recover the category label from the binary codes, the category information would have been encoded into the binary codes. Note that the category labels of some training images might be missing, to avoid the risk of misclassification of such images, we choose to simply ignore them in the classification task. Thus we define the classification loss of a single training image $X_i^{tr}$ as:
\begin{equation}
L^{cls}_i=-\sum_{c=1}^{C}\mathbb{I}\{y_i=c\}log\frac{s_c}{\sum_{l=1}^{C}{s_l}}
\label{eqn:cls}
\end{equation}
where the superscript $cls$ indicates classification, $\mathbb{I}\{\cdot\}$ is 1 when the condition is true and 0 otherwise, $s_l$ denotes the $l$-th output of the softmax classifier. For the case when $y_i=C+1$, namely, the category label of the $i$-th image is missing, for all $c\in\{1,\cdots,C\}$ we have $\mathbb{I}\{y_i=c\}=0$, thus the loss and gradient are both zeros, and those images without category labels will not contribute to the classification loss.

\subsection{Attributes Encoding}\label{sec:app:attr}

To preserve attribute similarity, the similar idea to Section \ref{sec:app:cls} is exploited, i.e. the attributes of images are encoded into the binary codes by applying a transformation that can recover the visual attributes from binary codes. Since the attributes are binary in this work, for each of the $m$ attributes, we define the loss as a logistic regression problem. To handle the missing label case, the standard formulation of logistic regression is modified to suit in our problem. Specifically, the $j$-th ($j\in\{1,2,\cdots,m\}$) attribute prediction loss of a single training image $X^{tr}_i$ is defined as a modified cross entropy loss:
\begin{equation}
L^{attr}_{ij}=-\mathbb{I}\{{\bf a}_{ij}\neq2\}[{\bf a}_{ij}log(p_{ij})+(1-{\bf a}_{ij})log(1-p_{ij})]
\label{eqn:attr}
\end{equation}
where the superscript $attr$ denotes attribute, $p_{ij}$ is the estimated probability that the $i$-th image possesses the $j$-th attribute.

Directly optimizing Eqn.(\ref{eqn:attr}) might lead to collapsed solution, since the distribution of some attributes are highly imbalanced (i.e. only a tiny portion of images have/do not have these attributes), even predicting all images as negative/positive would result in a relatively low loss. To alleviate the impact of sample imbalance, we propose a cost-sensitive version of Eqn.(\ref{eqn:attr}) instead:
\begin{equation}
\begin{aligned}
L^{attr}_{ij}({\bf w})=-\mathbb{I}\{{\bf a}_{ij}\neq2\}&[\frac{w_j}{w_j+1}{\bf a}_{ij}log(p_{ij})+\frac{1}{w_j+1}(1-{\bf a}_{ij})log(1-p_{ij})]
\end{aligned}
\label{eqn:attrweigh}
\end{equation}
where $w_j$ is a weighting parameter controlling the relative strength of the positive and negative samples. In practice, we set $w_j$ according to the ratio of the negative sample size to the positive sample size on the training set.

\subsection{Joint Optimization}

With the loss functions defined above, the CNN model can be trained with standard back propagation algorithm with mini-batches. However, directly adding up Eqn.(\ref{eqn:cls}) and Eqn.(\ref{eqn:attrweigh}) as the overall loss function may be problematic. To be specific, the values of Eqn.(\ref{eqn:cls}) and Eqn.(\ref{eqn:attrweigh}) might be in different orders of magnitudes. Moreover, due to missing labels, the loss corresponding to different attributes might also be in different orders of magnitudes. As a consequence, some parts of the loss might dominate and thus prevent the others from functioning. To tackle this problem, different parts of the loss function need to be scaled before added up. Suppose that in each iteration, the mini-batch consists of $n$ images, the overall loss function on a mini-batch is defined as follows:
\begin{equation}
L= \frac{1}{\sum_{t=1}^{n}\mathbb{I}\{y_t\leq C\}} \sum_{i=1}^nL^{cls}_i + \alpha\sum_{j=1}^m\sum_{i=1}^{n}\frac{L^{attr}_{ij}({\bf w})}{\sum_{t=1}^{n}\mathbb{I}\{{\bf a}_{tj}\neq2\}}
\label{eqn:overall}
\end{equation}
where $\alpha$ is an extra weighting parameter to control the relative strength of the classification loss and the attribute prediction loss. In case of $\sum_{t=1}^{n}\mathbb{I}\{y_t\leq C\}=0$ or $\sum_{t=1}^{n}\mathbb{I}\{{\bf a}_{tj}\neq2\}=0$, the corresponding loss term is set to zero.

The gradients of Eqn.(\ref{eqn:overall}) can be easily computed analogically to the standard softmax classifier, except for multiplying the weighting and scaling parameters, thus we do not bother to discuss them in detail. For the training images, their binary codes can be easily obtained by quantizing the corresponding binary-like network outputs.

\subsection{Retrieval}\label{sec:app:ret}

After the model is learned, the binary codes of new-coming images can be similarly obtained as above by propagating through the network and then quantizing the outputs of the binary-like layer. To accomplish the three retrieval tasks, we need to further recover the attribute predictions from the binary codes, which can be done by multiplying the binary codes with the attribute classifier weights. Note that the recovery of attribute prediction scores can be efficiently fulfilled by only a few summation operations, and only one more matrix (holding the attribute classifier weights) of size $k\times m$ (where $k$ is the code length, and $m$ is the number of attributes) needs to be stored compared to other hashing methods. Therefore, our method is efficient in both time and storage.

\section{Experiments}\label{sec:exp}

In this section, we extensively evaluate our method on two large-scale datasets. First we evaluated the impact of additional partially labelled data on the retrieval and attribute prediction tasks. Then the proposed DPH method was compared with the state-of-the-art retrieval methods on each of the three tasks to validate the advantages of our method.

\subsection{Experimental Settings}\label{sec:exp:setting}

\begin{figure}[t]
\setlength{\belowcaptionskip}{-0.1 cm}
\includegraphics[width=12.2 cm]{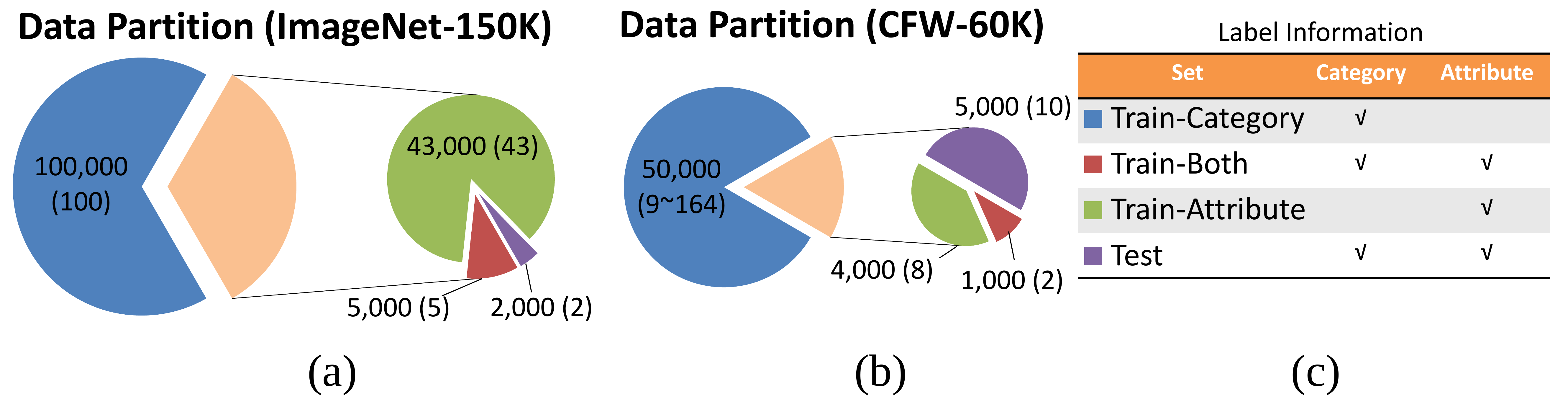}
\vspace{-0.8 cm}
\caption{Illustration of data partition in our experiments. (a) ImageNet-150K with 1,000 categories, (b) CFW-60K with 500 categories. The sizes of each set are presented in the figure, and the numbers in the brackets indicate the amount of images from each category. (c) The label information of the corresponding sets on the left, where the tick means the corresponding label is available. This figure is best viewed in color.}
\label{fig:data_partition}
\end{figure}

{\bf Datasets}: We evaluated our DPH method on two large-scale partially labelled datasets: (1) {\bf ImageNet-150K} is a subset of ILSVRC2012 dataset \cite{ILSVRC15} with 150,000 images. For each of the 1,000 categories, we selected 148 images from the training set and 2 images from the validation set. After that, 48 out of the 148 selected training images for each category and all the 2,000 selected validation images are manually annotated with 25 attributes (including color, texture, shape, material, and structure). We partitioned the dataset into 4 parts (Train-Category, Train-Both, Train-Attribute, and Test) as illustrated in Figure \ref{fig:data_partition}(a). Please refer to the supplementary materials for more details about this dataset. (2) {\bf CFW-60K} \cite{li2015two} is a purified subset of the original CFW dataset \cite{zhang2012finding} and contains 60,000 images of 500 subjects, among which 20 images of each subject (10K images in total) are annotated with 14 attributes. For the 10K images with attribute annotations, 5K (10 images for each subject) were used as Test set, and the rest 5K were further divided into two parts (Train-Both and Train-Attribute). For the remaining 50K images without attribute annotation, they were used as Train-Category set. The details of partitioning is illustrated in Figure \ref{fig:data_partition}(b). Please refer to the original publications \cite{zhang2012finding,li2015two} for more details about this dataset. On both datasets, the category labels of the Train-Attribute set were made unavailable to all methods.

{\bf Evaluation protocol}: All the evaluations are carried out solely on the Test set in a {\bf leave-one-out} manner, namely, each time we select one image from the Test set as query image, and the rest as database. The results are the mean performance over all images in the Test set. Since the three retrieval tasks are very different from each other, the details of the evaluation metrics for each task will be defined in their corresponding subsections (Section \ref{sec:exp:cat_ret}-\ref{sec:exp:comb_ret}) respectively.. 

{\bf Implementation details}: Although we have thousands of images as training data, our datasets are still relatively small in terms of training a CNN model from scratch. In consideration of generalization ability, the model parameters were initialized using pre-trained models. For ImageNet-150K, we used the publicly available CaffeNet model provided in the model zoo of Caffe \cite{jia2014caffe}. The model parameters from the conv1 layer to the fc7 layer were used to initialize our models. For CFW-60K, we adopted the CNN structure of \cite{yi2014learning} as the preceding layers (from conv1 to pool5). Since the pre-trained model is not available, we followed the original publication \cite{yi2014learning} to train the model with their published dataset, except for removing the contrastive loss for simplicity.

For ImageNet-150K, since the pre-trained model was trained on the same dataset as the target dataset, the model was trained for 40 epochs. In contrast, for CFW-60K, since the pre-trained model was obtained from a different dataset, the model was trained for 100 epochs. We set the learning rate to $10^{-3}$ for the preceding layers, and $10^{-2}$ for the other layers with a batch size of $200$. The momentum and weight decay parameters were set according to the original publications \cite{yi2014learning,jia2014caffe}. Besides, on both datasets, we empirically set the weighting parameter $\alpha=0.1$ in Eqn.(\ref{eqn:overall}). All the comparison deep learning methods were implemented with Caffe \cite{jia2014caffe} \footnote{The source code of DPH and the ImageNet-150K dataset will be released to the public.}.

\subsection{Evaluation of Partially Labelled Data}\label{sec:exp:module}

We first evaluate the impact of utilizing partially labelled data on both datasets by using 128-bit binary codes as example. For this purpose, 4 models were trained with different training sets: we name these models as Both ({\bf B}), Both + Attribute ({\bf B + A}), Both + Category ({\bf B + C}), and Both + Attribute + Category ({\bf B + A + C}) according to the data (please refer to Section \ref{sec:exp:setting} and Figure \ref{fig:data_partition} for details) used to train the specific model. In this subsection, the encoding of category and attributes are evaluated separately. For the category part, we rank the database images according to the Hamming distance between their binary codes and the codes of the query image, and the performance is measured by mAP, where images from the same category are deemed as relevant. For the attribute part, for simplicity, we calculate the F1-scores \cite{f1score} of predicting each attribute, and report the mean F1-score over all attributes. Note that since some attributes are highly unbalanced, e.g. most images do not possess the attribute ``orange'' in ImageNet-150K, F1-score can more faithfully reflect the real performance.

The comparison results are given in Table \ref{tab:module}. We can infer that: First, compared with the ``Both" model, exploiting extra training data (B + A and B + C) improves the performance of the corresponding task by a large margin. This observation can be explained by model overfitting, to be specific, in our experiments, when the model was trained only using the ``Train-Both'' set, the training loss approached zero while the test loss only decreased slightly. In contrast, when additional data was introduced to train the model, the training loss and test loss of the corresponding tasks were always on the same scale as normally expected. This justifies our motivation of using partially labelled data to train the CNN models to alleviate overfitting. Second, compared with training solely on ``Train-Both'' set, using both kinds of additional data can significantly improve the performance on both tasks (the row ``B + A + C'' in Table \ref{tab:module}), and the performance of this dual-purpose model is comparable with or even better than the performances of the ``B + A'' and ``B + C'' models, confirming that it is feasible to simultaneously embed category and visual attributes into the binary codes by exploiting partially labelled data. In the following experiments, all our models are trained with the ``B + A + C'' setting.

\begin{table}[t]
\begin{small}
\setlength{\belowcaptionskip}{0.1 cm}
\caption{Comparison of the 128-bit models trained with different combinations of training data. The retrieval mAP and mean F1-score over all attributes are shown in the last two columns respectively. B: Both, A: Attribute, C: Category.}
\label{tab:module}
\begin{center}
\begin{tabular}{c||c|c|c||c|c|c}
\hline
Model & Dataset & mAP & mean F1-score & Dataset & mAP & mean F1-score\\
\hline
\begin{tabular}{c}
B\\
B + A\\
B + C\\
B + A + C\\
\end{tabular} &
\begin{tabular}{c}
ImageNet-\\
150K
\end{tabular}&
\begin{tabular}{c}
0.248\\
0.239\\
0.336\\
0.343\\
\end{tabular} &
\begin{tabular}{c}
0.753\\
0.856\\
0.828\\
0.879
\end{tabular} &
\begin{tabular}{c}
CFW-\\
60K
\end{tabular} &
\begin{tabular}{c}
0.095\\
0.088\\
0.233\\
0.241\\
\end{tabular} &
\begin{tabular}{c}
0.817\\
0.867\\
0.814\\
0.877
\end{tabular}\\
\hline
\end{tabular}
\end{center}
\end{small}
\vspace{-0.8 cm}
\end{table}

\subsection{Evaluation of Category Retrieval}\label{sec:exp:cat_ret}

In this subsection, we test the effectiveness of our DPH method on the first task specified in Section \ref{sec:intro}, namely, given a query image, retrieving images of the same category from the database. The retrieval is done by ranking the binary codes of database images according to the Hamming distances to the query image.

{\bf Comparative methods}: We compare with seven representative hashing methods: ITQ \cite{gong2011iterative}, CCA-ITQ \cite{gong2011iterative}, DBC \cite{rastegari2012attribute}, KSH \cite{liu2012supervised}, SDH \cite{shen2015supervised}, DNNH \cite{lai2015dnnh}, and DLBHC \cite{lin2015deep}, including representative linear and non-linear conventional hashing methods as well as the state-of-the-art deep CNN-based methods. 

For fair comparison, the ``shallow'' hashing methods were trained using the L2-normalized CNN features extracted from the pre-trained CNN models (described in Section \ref{sec:exp:setting}). The comparative methods were implemented using the source code provided by the original authors except for LSH. Instead, the projection parameters of LSH were randomly drawn from a normal distribution. As for the ``deep'' methods, DLBHC and DNNH exploited the same preceding layers as our DPH method, and were initialized with the identical pre-trained models as ours. In particular, since the category number is larger than the batch size in our setting, DNNH would fail to converge if the training images are randomly shuffled, mainly because the number of ``valid'' triplets in each iteration is too small. To make DNNH converge successfully, we hence randomly selected 10 categories and 20 images per category to form each mini-batch.

The comparative methods were trained using the combination of the two sets ``Train-Both'' and ``Train-Category'' to preserve the category similarity. Since KSH demands large amount of memory to store the kernel matrix ($O(N^2)$, where $N$ is the number of training images), we used 20,000 images randomly selected from the training set for this method, which has already consumed more than 16GB of memory in the training stage. All the hyper-parameters of the comparative methods were tuned carefully according to the original publications. The experiments were carried on $\{16,32,64,128,256\}$-bit binary codes. 

{\bf Evaluation metric}: For evaluation, we use mean Average Precision (mAP), where images with the same category label are considered as relevant.

\begin{table}[t]
\setlength{\belowcaptionskip}{-0.0 cm}
\caption{Comparison of category retrieval performance (mAP) of our method and other comparative hashing methods on ImageNet-150K and CFW-60K. The best performance of each code length is highlighted in boldface.}
\label{tab:cat_ret}
\begin{small}
\begin{center}
\begin{tabular}{c|ccccc|ccccc}
\hline
& \multicolumn{5}{c|}{ImageNet-150K} & \multicolumn{5}{c}{CFW-60K}\\
& 16-bit & 32-bit & 64-bit & 128-bit & 256-bit & 16-bit & 32-bit & 64-bit & 128-bit & 256-bit\\
\hline
ITQ \cite{gong2011iterative} & 0.102 & 0.167 &0.235 & 0.284 & 0.310  & 0.039 & 0.058 & 0.079 & 0.112 & 0.135 \\
CCA-ITQ \cite{gong2011iterative} & 0.090 & 0.157 & 0.223 & 0.294 & 0.341 & 0.048 & 0.069 & 0.090 & 0.113 & 0.140 \\
DBC \cite{rastegari2012attribute} & 0.207 & 0.264 & 0.308 & 0.344 & 0.369 & 0.045 & 0.060 & 0.072 & 0.099 & 0.129 \\
KSH \cite{liu2012supervised} & 0.110 & 0.181 & 0.253 & 0.293 & 0.320 & 0.046 & 0.063 & 0.086 & 0.111 & 0.117 \\
SDH \cite{shen2015supervised} & 0.082 & 0.143 & 0.222 & 0.288 & 0.322 & 0.026 & 0.049 & 0.095 & 0.140 & 0.183 \\
DNNH \cite{lai2015dnnh} & 0.102 & 0.147 & 0.213 & 0.267 & 0.298 & 0.035 & 0.058 & 0.100 & 0.148 & 0.185 \\
DLBHC \cite{lin2015deep} & 0.197 & 0.263 & 0.310 & 0.339 & 0.357 & 0.068 & 0.109 & 0.173 & 0.235 & 0.279 \\
\hline
DPH & 0.212 & 0.274 & 0.322 & 0.343 & 0.353 & 0.064 & 0.112 & 0.186 & 0.241 & 0.274 \\
\hline
\end{tabular}
\end{center}
\end{small}
\vspace{-0.8 cm}
\end{table}

{\bf Results}: The comparison results are shown in Table \ref{tab:cat_ret}. We have the following observations: {\bf First},  when equipped with CNN features, the conventional non-linear hashing method KSH can hardly improve the retrieval performance over linear methods. One possible explanation is that the CNN has mapped the images to a feature space where images from different categories are roughly linearly separable, thus KSH can hardly benefit from the non-linearity of kernel space. In addition, the smaller training set of KSH is also a possible explanation. {\bf Second}, in terms of retrieval mAP, CNN-based methods significantly improve over conventional hashing methods on CFW-60K, yet have marginal improvement on ImageNet-150K. Note that the pre-trained model on CFW-60K was obtained from a different dataset, while on ImageNet-150K from the same one, validating the advantage of CNN-based hashing methods lies in learning image representations which are more suitable for the data at hand than pre-defined features. {\bf Third}, DNNH performs relatively worse than the other two CNN-based methods \footnote{The source code of DNNH was provided by the original authors, and our re-implementation on NUS-WIDE achieved similar result as reported in \cite{lai2015dnnh}}. While this might be attributed to the batch generation scheme particularly designed for this method as described above, it seems to imply that the training data should be carefully organized for DNNH to yield favorable performance. {\bf Fourth}, the performance of DPH is among the top of all methods, even though the binary codes were learned for jointly tackling two kinds of different tasks, indicating that our dual purpose hash codes is competent to fulfil the first individual task - category retrieval.

\subsection{Evaluation of Attribute Retrieval}\label{sec:exp:attr_ret}

In this subsection, we test the effectiveness of our DPH method on the second task described in Section \ref{sec:intro}. The attribute prediction scores of DPH can be recovered from the binary codes using the method described in Section \ref{sec:app:ret}. In this experiment, given an image, we randomly select at most three attributes as query, whose values are specified by the image (thus can be either positive or negative). The system is required to retrieve images such that the selected attributes of the top ranked images are the same as the ones of the query image. To be specific, the database images were ranked in descending order by the products of attribute prediction scores. 

{\bf Comparative methods}: We compare with three baseline methods for attribute prediction: 1) Similar to \cite{kumar2008facetracer}, we train linear SVM classifiers to predict attributes (in the experiments, we found that replacing the linear SVMs with kernel SVMs only gives marginal improvement, thus we adopted the linear SVMs for efficiency), using the same CNN features as the "shallow" hashing methods described in Section \ref{sec:exp:cat_ret}. Then the prediction scores are normalized to the range of $(0,1)$ using $sigmoid$ function. We denote this method as {\bf SVM-real}, where ``real'' indicates that the models were trained on real-valued features. 2) We replace the CNN features in SVM-real with the 256-bit binary codes produced by DLBHC in Section \ref{sec:exp:cat_ret}. This baseline is used to evaluate the necessity of jointly encoding the category and attributes. We denote this method as {\bf SVM-binary}. 3) We finetune the pretrained CNN models to solely predict the attributes. For this purpose, we modified our network structure by removing both the binary-like layer and the classification loss, and concatenating the attribute prediction loss right after the preceding layers. The models were trained using the combination of the two sets ``Train-Both'' and ``Train-Attribute'', and the hyper-parameters were set as described in Section \ref{sec:exp:setting}. We denote this method as {\bf CNN-attribute}.

{\bf Evaluation metric}: In this task, we also use mAP to measure the retrieval performance. Images that match with the query image at all selected attributes are considered as relevant. Note that in this experiment, the predicted attributes of all images (both query and database images) were used for retrieval, while the evaluation is performed on the ground-truth attribute labels. As a result, both wrong predictions of the query image and the database images would hurt the performance. We report the average mAP over all valid attribute queries.

\begin{figure}[t]
\setlength{\belowcaptionskip}{-0.5 cm}
\begin{center}
\includegraphics[width=12.0cm]{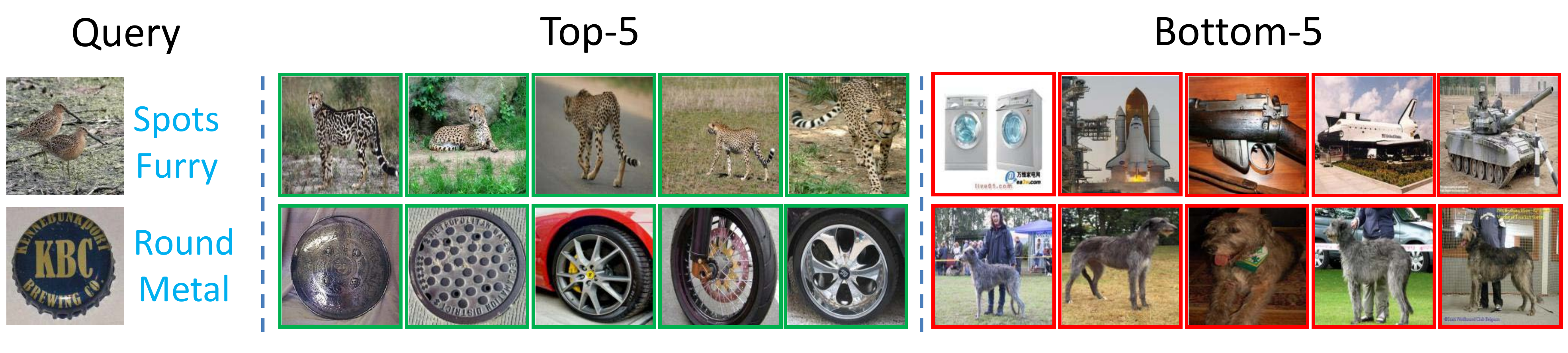}\\
(a)\\
\includegraphics[width=12.0cm]{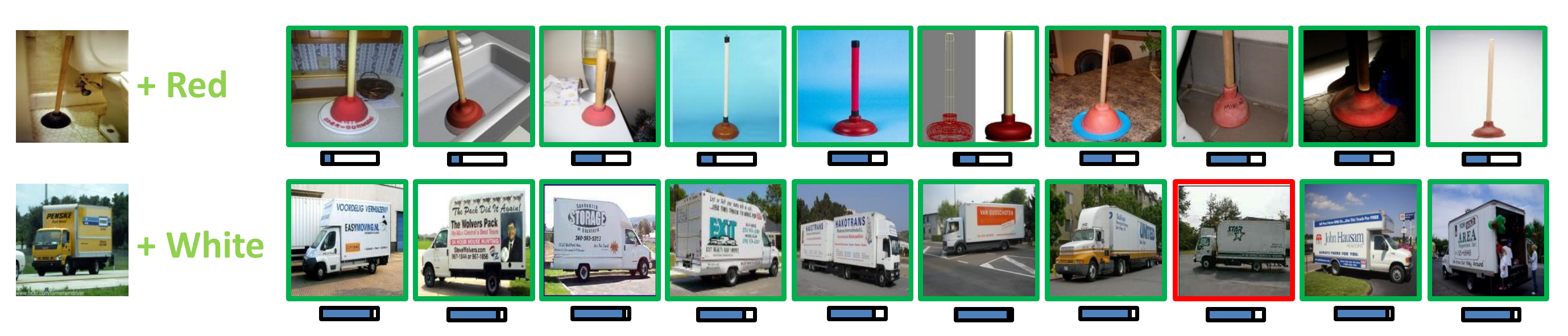}\\
(b)
\end{center}
\vspace{-0.6 cm}
\caption{Some real retrieval cases of the two attribute-oriented tasks on ImageNet-150K. In these tasks, images in the ``Test'' set were used as queries, and the ``Train-Both'' set and ``Train-Attribute'' set were used together as database. (a) The two rows are from task \uppercase\expandafter{\romannumeral2}, and the interested attributes are listed on the right side of the query image. (b) The two rows are from task \uppercase\expandafter{\romannumeral3}, and the top-10 ranked images are displayed. The notations here are consistent with Figure \ref{fig:illustration}(b). This figure is best viewed in color.}
\label{fig:real_case}
\vspace{-0.1 cm}
\end{figure}

{\bf Results}: The results are given in Table \ref{tab:map_attribute}, note that SVM-real and CNN-attribute have very close performances on ImageNet-150K, and their curves overlap. On both datasets, our 256-bit binary codes achieves comparable or even better performance than the baseline methods. Our method does not need to store the real-valued attribute prediction scores, thus compared to SVM-real and CNN-attribute, the storage space required by our method is much smaller. SVM-binary achieves similar performance with our method on ImageNet-150K, but much worse on CFW-60K. This could possibly be explained by the fact that ImageNet-150K contains more categories and attributes than CFW-60K, and the variation is thus more complex. As a result, the 256-bit code might be too short for this task. From the tendency in Figure \ref{fig:map_attribute}(a), we can hopefully expect that longer codes of our DPH method could achieve better performance. Some real retrieval results on this task are provided in Figure \ref{fig:real_case}(a). Please refer to the supplementary materials for more examples.

\begin{table}[t]
\caption{Comparison of attribute retrieval performance (average mAP) of our method and other comparative methods on (a) ImageNet-150K and (b) CFW-60K. Note that SVM-real and CNN-attribute do not use binary code as features, thus their performance do not vary with code lengths.}
\label{tab:map_attribute}
\begin{small}
\begin{center}
\begin{tabular}{c|ccccc|ccccc}
\hline
& \multicolumn{5}{c|}{ImageNet-150K} & \multicolumn{5}{c}{CFW-60K}\\
& 16-bit & 32-bit & 64-bit & 128-bit & 256-bit & 16-bit & 32-bit & 64-bit & 128-bit & 256-bit\\
\hline
SVM-real & \multicolumn{5}{c|}{0.903} & \multicolumn{5}{c}{0.765}\\
CNN-attribute & \multicolumn{5}{c|}{0.902} & \multicolumn{5}{c}{0.771}\\
SVM-binary & 0.805 & 0.823 & 0.844 & 0.861 & 0.871 & 0.661 & 0.680 & 0.693 & 0.711 & 0.729\\
\hline
DPH & 0.806 & 0.828 & 0.842 & 0.859 & 0.868 & 0.695 & 0.726 & 0.758 & 0.785 & 0.804\\
\hline
\end{tabular}
\end{center}
\end{small}
\vspace{-0.8cm}
\end{table}

\subsection{Evaluation of Combined Retrieval}\label{sec:exp:comb_ret}

In this subsection, we evaluate our DPH method on the third retrieval task described in Section \ref{sec:intro}. In this experiment, the system is required to retrieve images belonging to the same category as the query image, while possessing an attribute that is absent in the query image. To accomplish this task, we use the attribute predictions to filter out the images that do not match in terms of the specified attribute, and then rank the remaining images using the Hamming distances. We compare the results of using 256-bit binary codes in this experiment.

{\bf Comparative methods}: Since this is a relatively new task, we compare our DPH with two methods: 1) {\bf JLBC} \cite{li2015two} is trained on ``Train-Both'' set, since it can only accept fully annotated images as training inputs. We used the same CNN features as described above to train this method. 2) A combination of DLBHC \cite{lin2015deep} and the CNN-attribute model in Section \ref{sec:exp:attr_ret} (CNN-attribute for attribute prediction and DLBHC for Hamming distance ranking, which corresponds to training two separate models). The DLBHC model used here was trained to produce $(256 - m)$-bit binary codes, where $m$ is the number of attributes, and the predictions of CNN-attribute were quantized to binary, thus the storage cost of this method is equal to our DPH method. We denote this method as {\bf Multiple-model}.

{\bf Evaluation metric}: Similar to the previous sections, the query attribute is acquired by attribute predictors, and the performance is evaluated using the ground-truth labels. Only images that match the query image in terms of category and possess the query attribute are considered as relevant. We use $recall@\{5,10,20,50,75,100\}$ to evaluate the different methods. In case that the database does not contain any true matches, the recall of such query is simply ignored. We report the average recall over all valid queries.

\begin{figure}[t]
\setlength{\belowcaptionskip}{-0.5 cm}
\begin{center}
\begin{tabular}{cc}
\includegraphics[width=5.5 cm]{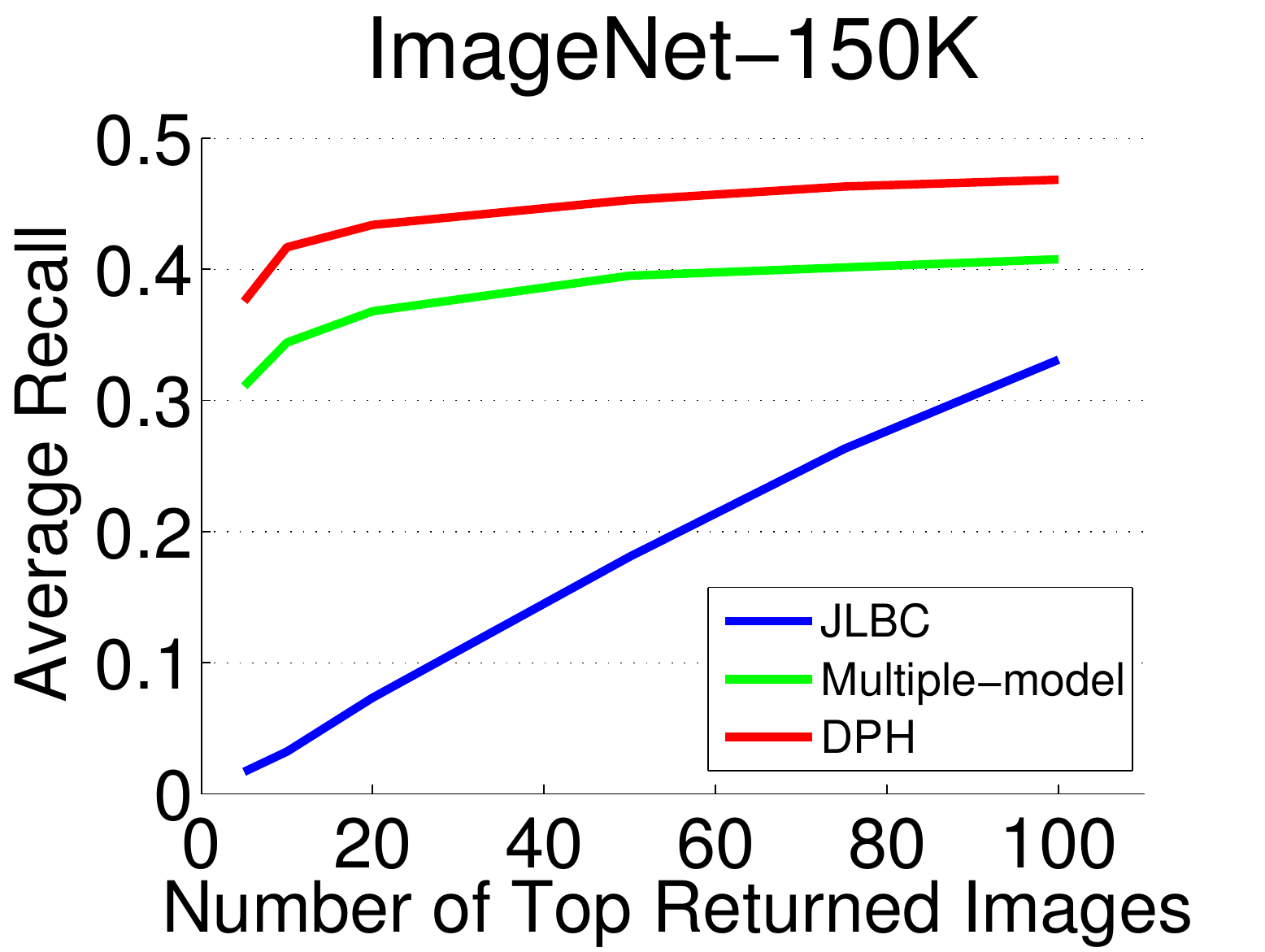} &
\includegraphics[width=5.5 cm]{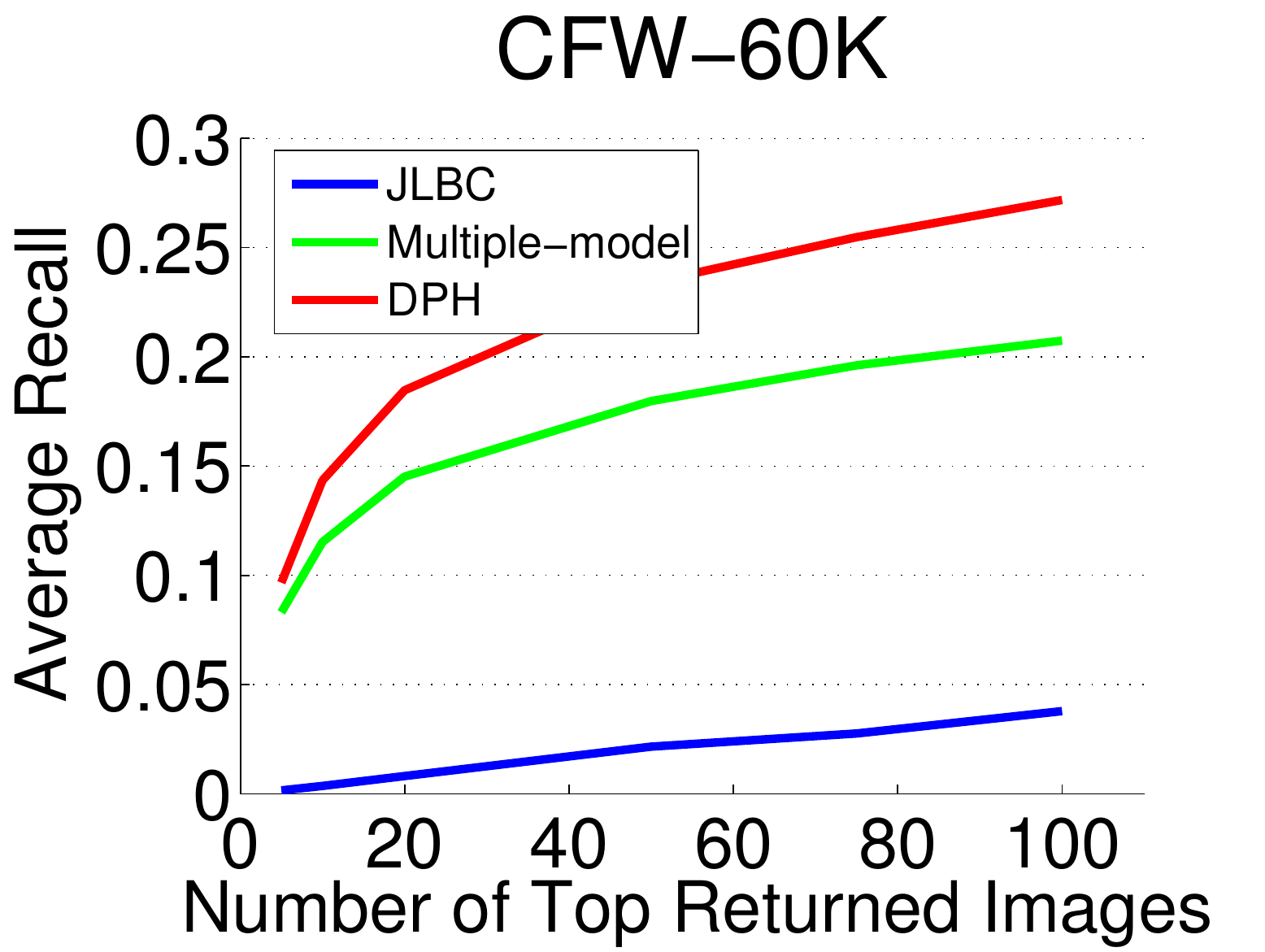} \\
(a) & (b)
\end{tabular}
\end{center}
\vspace{-0.4 cm}
\caption{Comparison of combined retrieval performance (average recall) of our method and other comparative methods on (a) ImageNet-150K and (b) CFW-60K. The results were obtained by 256-bit binary code.}
\label{fig:recall_combined}
\end{figure}

{\bf Results}: The results are shown in Figure \ref{fig:recall_combined}. Our method consistently outperforms the comparative methods. The performance of JLBC on CFW-60K is very unsatisfactory, even though CNN features was used to train this model. This result confirms that our end-to-end framework is necessary for learning dual purpose hash codes. Although each model of the "Multiple-model" method performs quite well on its own task, their combination is clearly outperformed by our DPH method. A possible explanation is that the codes learned by these two models are redundant, while our DPH can suppress the redundancy between category and attributes by exploiting the correlation between them, thus the total amount of information they actually carry is less than our dual purpose codes. Moreover, the Multiple-model method needs two networks to produce the binary codes, thus the computation cost is twice as much as our method. We provide some real retrieval results on this task in Figure \ref{fig:real_case}(b). Please refer to the supplementary materials for more results.

\vspace{-0.25 cm}

\subsection{Discussion}


To sum up, our DPH method utilized more supervised information than those state-of-the-art methods specifically designed for each individual task (i.e. category retrieval and attribute retrieval), one thus expects that DPH should naturally yield better performances. Indeed, since some attributes often vary significantly even within a single class (e.g. color attributes of towels), the additional attribute information might even make the learning of category more difficult. Even though, the performances of our binary codes on the three retrieval tasks are still satisfactory, while the computation cost of our method is much lower than training multiple models, indicating that jointly preserving both category and attribute similarities for the three tasks is advantageous.


\vspace{-0.25 cm}

\section{Conclusions}\label{sec:con}

In this paper we propose a method to learn hash functions that simultaneously preserve category and attribute similarities for multiple retrieval tasks. Our DPH method has achieved very competitive retrieval performances against state-of-the-art methods specifically designed for each individual task. The promising performance of our method can be attributed to: a) The utilization of CNN models for hierarchically capturing correlation between category and attributes in an end-to-end manner. b) The loss functions specifically designed for the partially labelled training data, which can significantly improve the generalization ability of the models. Note that our framework is quite general, thus more powerful network structures and loss functions can be easily incorporated to further improve the performance of our method. 

\end{spacing}

\clearpage

\bibliographystyle{splncs}
\bibliography{DPH_arXiv}

\clearpage

\title{Supplementary Materials: Dual Purpose Hashing }
\author{}
\institute{}

\maketitle

The following sections give details about the attributes defined on ImageNet-150K dataset, and additional real retrieval results. This material is best viewed in color.

\section{Example Images of Attributes}

\begin{figure}
\begin{center}
\includegraphics[width=12.0 cm]{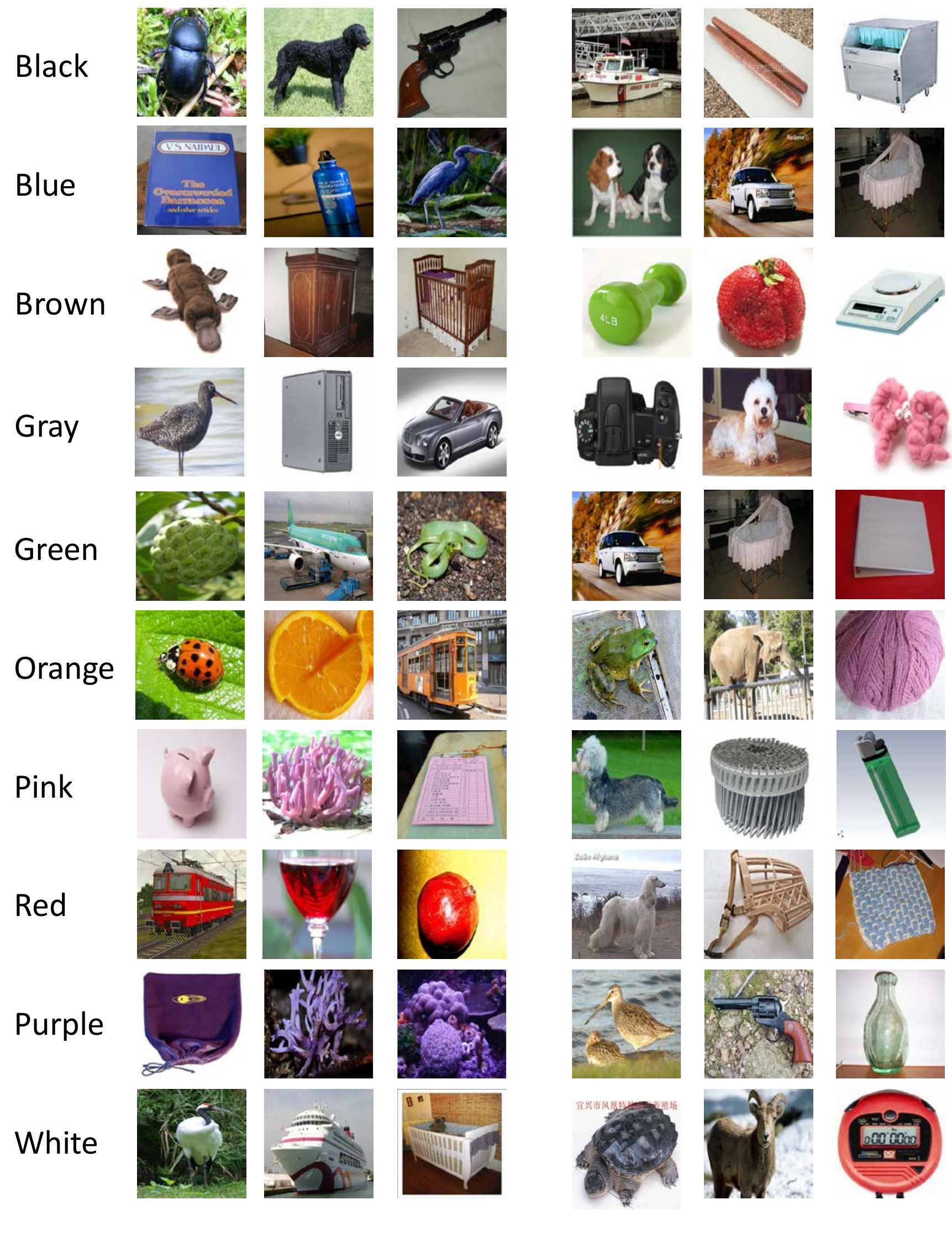}
\end{center}
\caption{Example images of attributes on ImageNet-150K. For each attribute, three positive samples (the leftmost three) and three negative samples (the rightmost three) are shown in this figure.}
\label{fig:ImageNet_attr_1}
\end{figure}

\begin{figure}
\begin{center}
\includegraphics[width=12.0 cm]{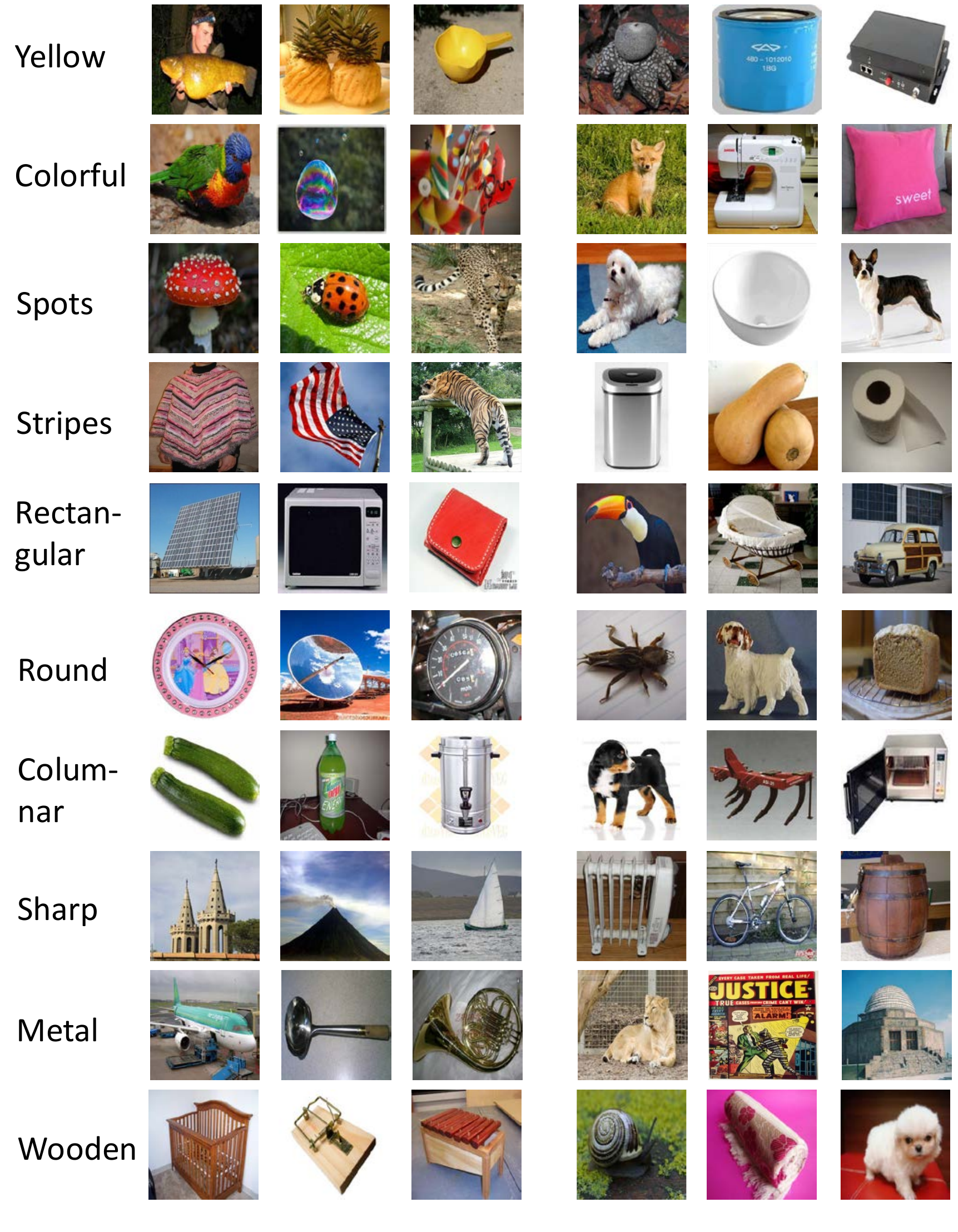}
\end{center}
\caption{Example images of attributes on ImageNet-150K. For each attribute, three positive samples (the leftmost three) and three negative samples (the rightmost three) are shown in this figure.}
\label{fig:ImageNet_attr_2}
\end{figure}

\begin{figure}
\begin{center}
\includegraphics[width=12.0 cm]{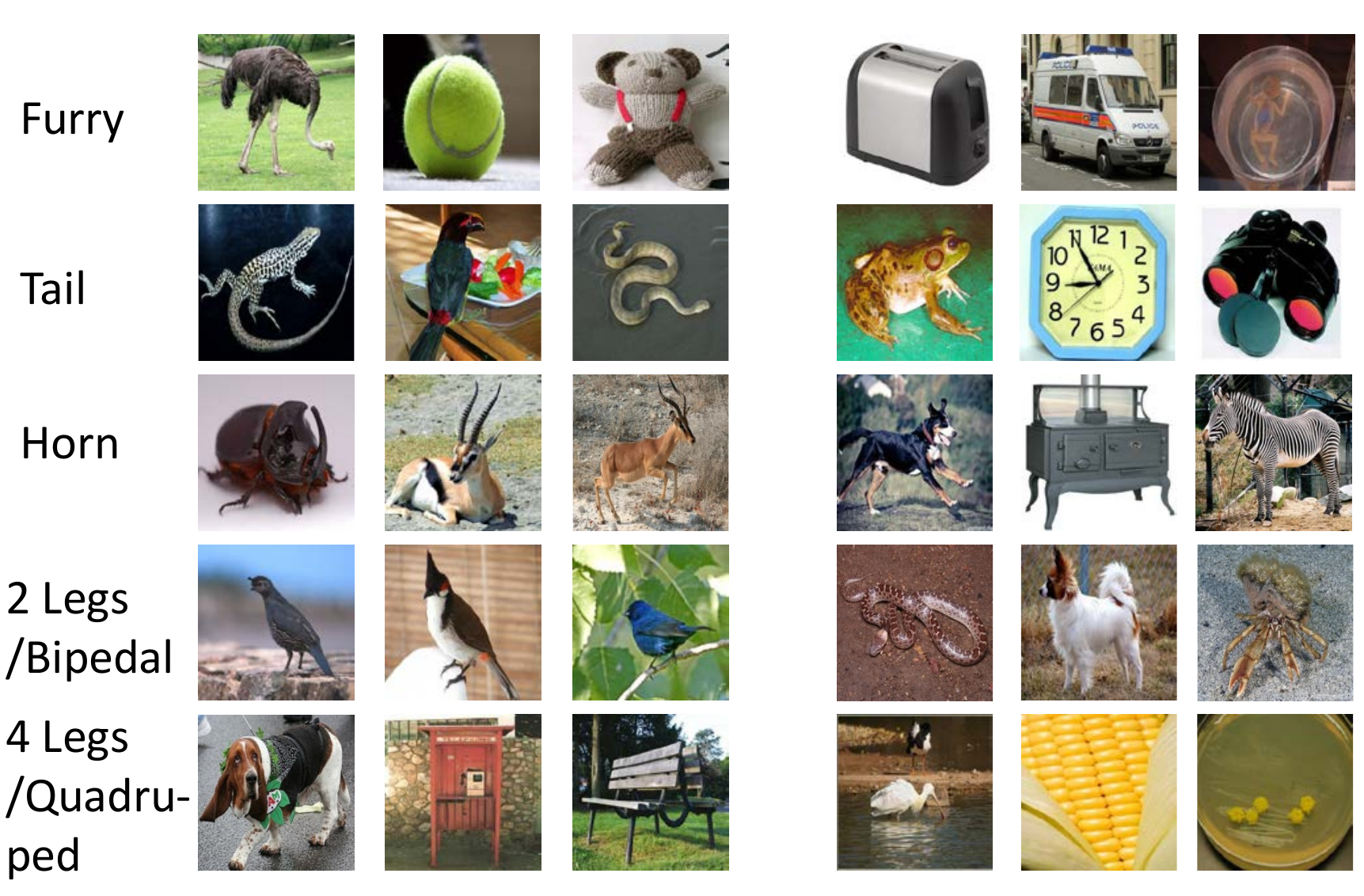}
\end{center}
\caption{Example images of attributes on ImageNet-150K. For each attribute, three positive samples (the leftmost three) and three negative samples (the rightmost three) are shown in this figure.}
\label{fig:ImageNet_attr_3}
\end{figure}

In this section, we provide example images of each attribute defined on ImageNet-150K (25 attributes, including color, texture, shape, material, and structure). The attributes were defined and annotated mainly based on the ImageNet-attribute [S1] and Animals with Attribute (AwA) [S2] datasets. Compared to [S1], our dataset covers much more categories (1000 vs 384) and images (50,000 vs 9,600). For each attribute, three positive samples along with three negative samples are shown in Figure \ref{fig:ImageNet_attr_1}, \ref{fig:ImageNet_attr_2}, and \ref{fig:ImageNet_attr_3} (the leftmost three in each row are positive samples and the rest are negative samples). In our experiments, the attributes are binary, namely, an image either has or does not have the attribute.

\section{Real Retrieval Cases}

This section gives more real retrieval cases on the attribute-oriented retrieval tasks described in Sections 4.4 and 4.5 of the main paper(the results were obtained with 256-bit binary codes).

\subsection{Results on CFW-60K}

\begin{figure}
\begin{center}
\begin{tabular}{c}
\includegraphics[width = 12.0 cm]{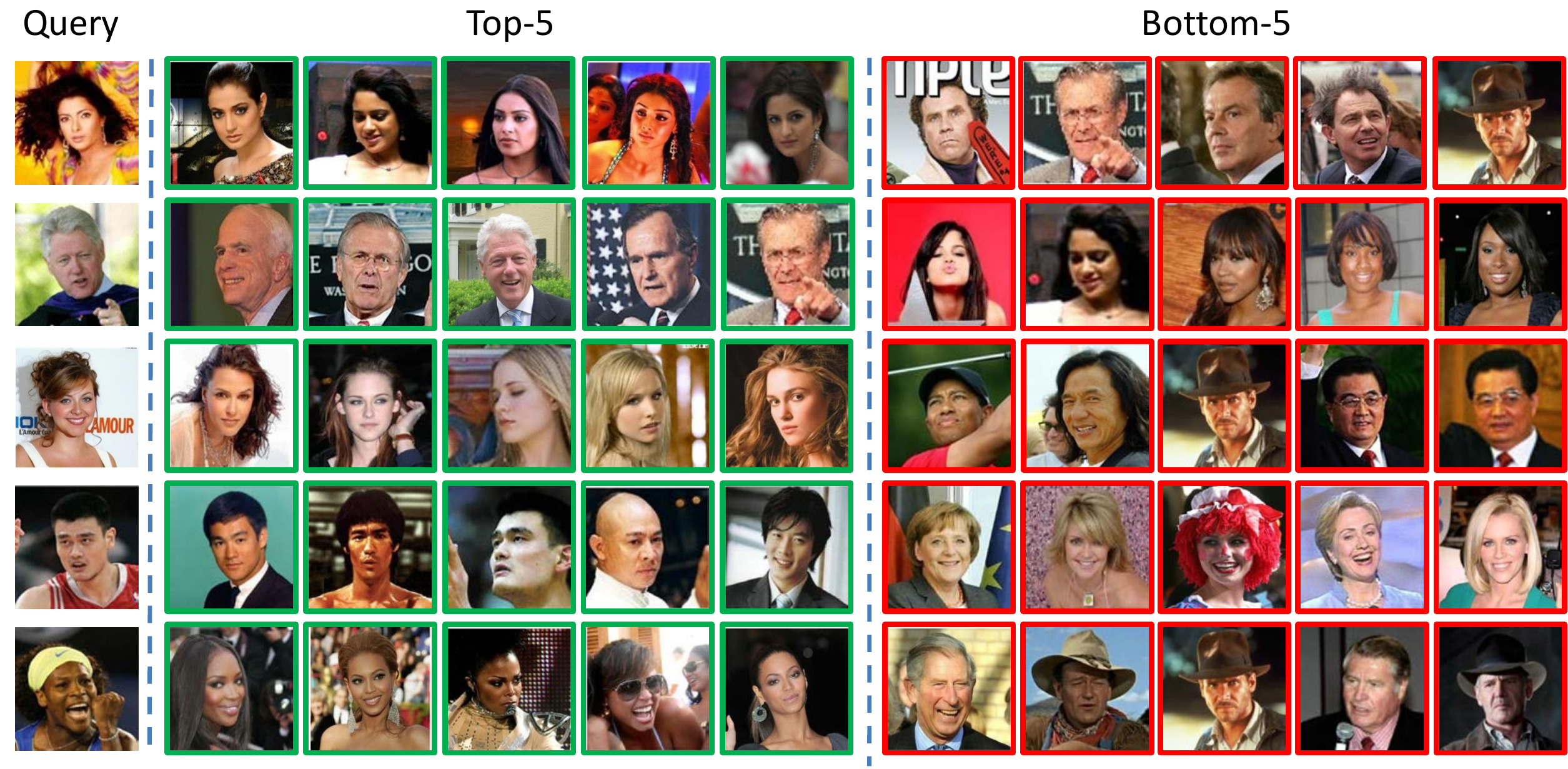}\\
(a)\\
\includegraphics[width = 12.0 cm]{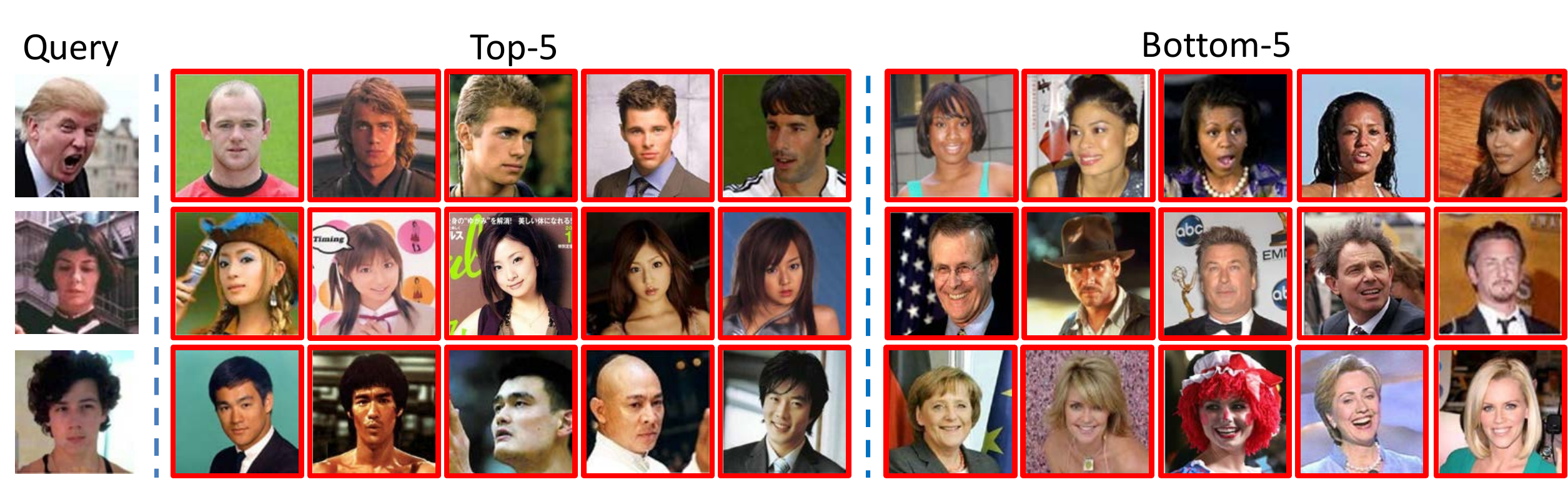}\\
(b)\\
\end{tabular}
\end{center}
\caption{Some real retrieval results of task \uppercase\expandafter{\romannumeral2} on CFW-60K (retrieving images of subjects with the same gender, race, and age group as the subject in the query image). The results were obtained with 256-bit binary codes. The notations are consistent with the main paper (please refer to Figure 1 in the main paper for details). (a) successful cases, (b) failed cases. In the failed cases, the predicted gender, race, and age group of the 3 query images are: 1) male + white + young (groundtruth: male + white + mid-aged), 2) female + Asian + young (groundtruth: female + white + young), 3) male + Asian + young (groundtruth: male + white + young). Note that as mentioned in our main paper, in this task II the predicted attributes of all images (both query and database images) were used for retrieval, while the evaluation is performed on the ground-truth attribute labels. As a result, both wrong predictions of the query image and the database images would cause a mismatch.}
\label{fig:CFW_task1}
\end{figure}

\begin{figure}
\begin{center}
\includegraphics[width = 12.0 cm]{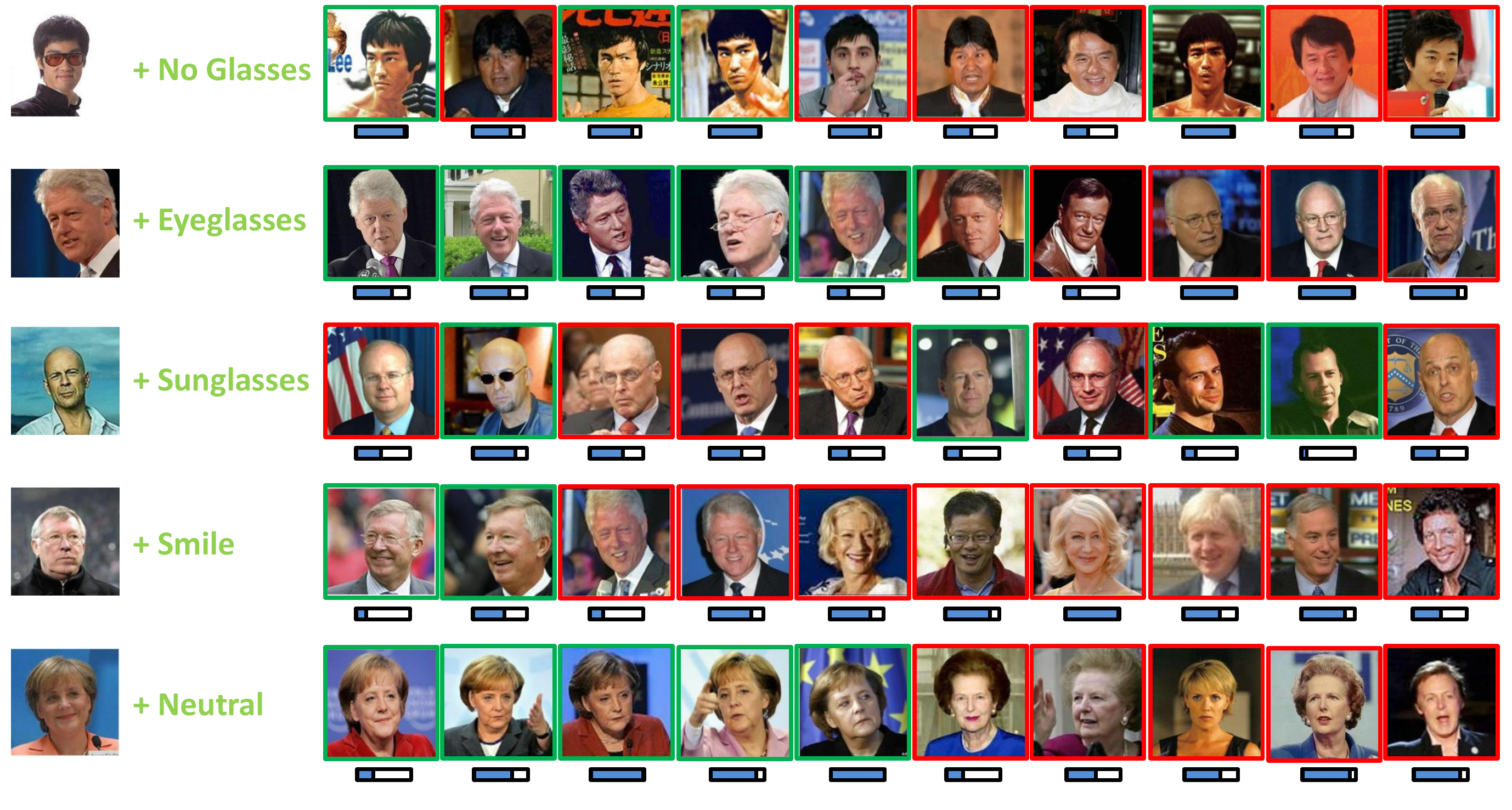}
\end{center}
\caption{Some real retrieval results of task \uppercase\expandafter{\romannumeral3} on CFW-60K. The notations are consistent with the main paper (please refer to Figure 1 in the main paper for details).}
\label{fig:CFW_task3}
\end{figure}

The results of task \uppercase\expandafter{\romannumeral2} and task \uppercase\expandafter{\romannumeral3} on CFW-60K are shown in Figure \ref{fig:CFW_task1} and \ref{fig:CFW_task3} respectively. In task \uppercase\expandafter{\romannumeral2}, the system is required to retrieve images of subjects with the same gender, race, and age group as the subject in the query image. As we can see from the failed cases (Figure \ref{fig:CFW_task1}(b)), for each query image, though the top feedbacks fail to match the exact attributes of the query, all of them have the same gender, race, and age group. By further investigating the failed cases, we found that the main cause is the incorrect attribute predictions of the query image. In task \uppercase\expandafter{\romannumeral3}, either inaccurate attribute prediction or the incapability of binary codes in preserving category similarity would result in failed cases. Here we only show the successful cases to demonstrate the potential of our method in this challenging realistic retrieval scenario.

\subsection{Results on ImageNet-150K}

\begin{figure}
\begin{center}
\includegraphics[width = 12.0 cm]{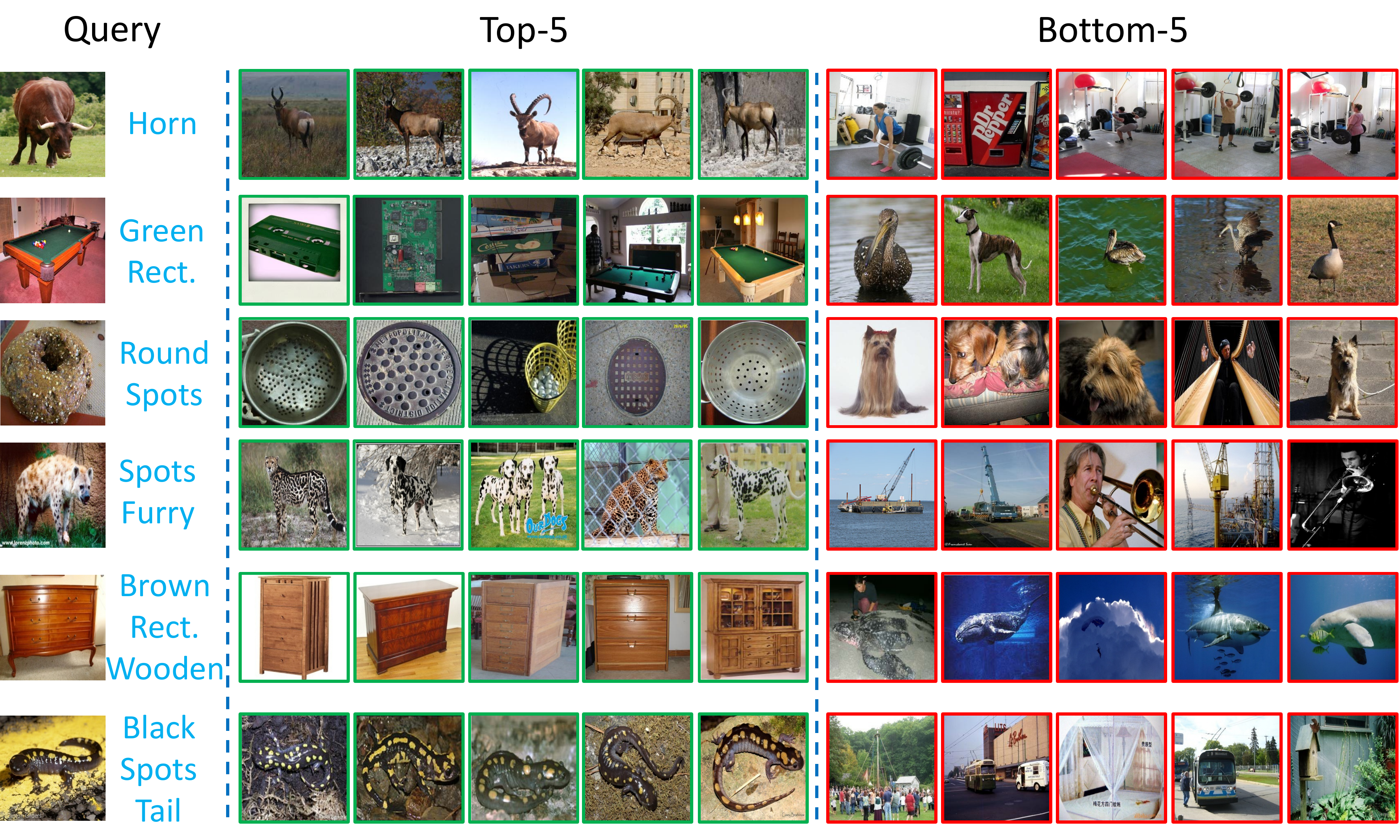}
\end{center}
\caption{Some real retrieval results of task \uppercase\expandafter{\romannumeral2} on ImageNet-150K, the interested attributes are listed on the right side of the query images, including color, texture, shape, material, and structure. Images in the ``Test'' set were used as queries, and the ``Train-Both'' set and ``Train-Attribute'' set were used as database. The results were obtained with 256-bit binary codes. The notations are consistent with the main paper (please refer to Figure 1 in the main paper for details).}
\label{fig:ImageNet_task1}
\end{figure}

\begin{figure}
\begin{center}
\includegraphics[width = 12.0 cm]{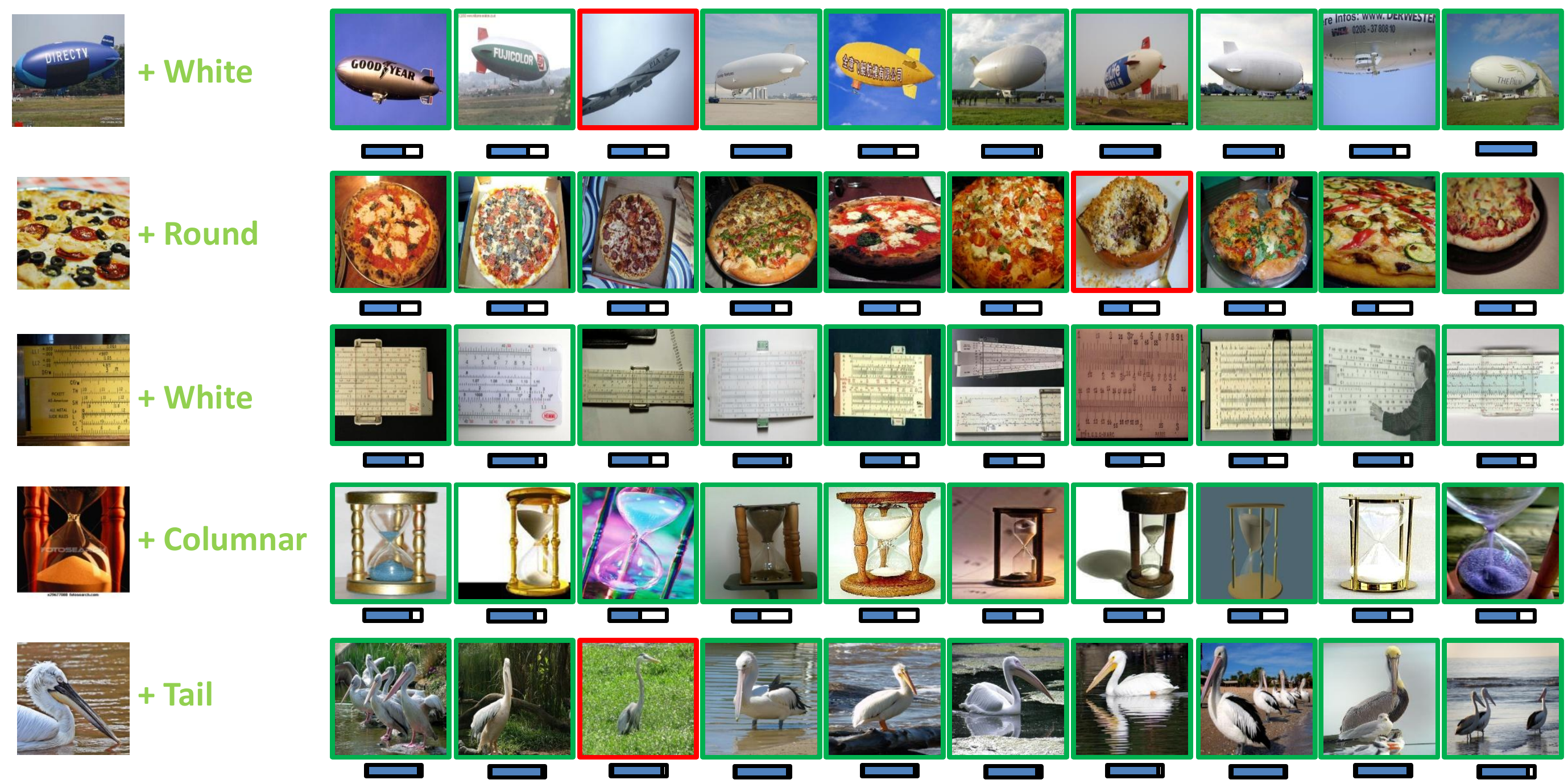}
\end{center}
\caption{Some real retrieval results of task \uppercase\expandafter{\romannumeral3} on ImageNet-150K. Images in the ``Test'' set was used as queries, and the ``Train-Both'' set and ``Train-Attribute'' set were used as database. The notations are consistent with the main paper (please refer to Figure 1 in the main paper for details). Note that for each of the second and fifth query, one of the top-10 feedbacks (that is bounded by red box) does not match the query in terms of category (the ground-truth category of the queries are ``moving van'' and ``pelican'' respectively, while the ground-truth category of the wrong feedbacks are ``trailer truck'' and ``crane'' respectively). We can see that the wrong feedbacks look very similar to the query images, even humans would have some difficulty to perform such a fine-grained categorization task, thus it is reasonable that the retrieval system made such a mistake.}
\label{fig:ImageNet_task3}
\end{figure}

The results on ImageNet-150K are shown in Figure \ref{fig:ImageNet_task1} and \ref{fig:ImageNet_task3}. For this dataset, since there are only two images from each category in the ``Test'' set, to better evaluate our method for qualitative demonstration, in this supplemental experiment we used the ``Test'' set as query images, and retrieved images from both ``Train-Both'' and ``Train-Attribute'' sets. Some successful retrieval results on task \uppercase\expandafter{\romannumeral2} and task \uppercase\expandafter{\romannumeral3} are provided, suggesting that our method has the potential to be applied in these two realistic yet very challenging object retrieval scenarios.
\vspace{0.5cm}
\\
{\bf References}
\\\
\\\
[S1] Russakovsky, O., Fei-Fei, L.: Attribute learning in large-scale datasets. In: ECCV workshop. Springer (2010) 1-14
\\\
\\\
[S2] Lampert, C.H., Nickisch, H., Harmeling, S.: Learning to detect unseen object classes by between-class attribute transfer. In: Computer Vision and Pattern Recognition, 2009. CVPR 2009. IEEE Conference on, IEEE (2009) 951-958

\end{document}